\pdfoutput=1
\documentclass[11pt]{article}

\usepackage{acl}

\usepackage{times}
\usepackage{latexsym}

\usepackage[T1]{fontenc}
\usepackage[utf8]{inputenc}

\usepackage{microtype}
\usepackage{algorithm}
\usepackage{algorithmic}
\usepackage{graphicx}
\usepackage{subfig}
\usepackage{float}
\usepackage{amsthm,amsmath,amssymb}
\usepackage{mathrsfs}
\usepackage{multirow}
\usepackage{booktabs}
\usepackage{threeparttable}
\usepackage{hyperref}
\usepackage{longtable}
\usepackage{color}
\title{OneNet: A Fine-Tuning Free Framework for Few-Shot Entity Linking \\ via Large Language Model Prompting}

\author{Xukai Liu, Ye Liu, Kai Zhang\thanks{~~corresponding author.}, Kehang Wang, Qi Liu, Enhong Chen \\
 State Key Laboratory of Cognitive Intelligence, University of Science and Technology of China \\
  \normalsize \texttt{\{chthollylxk,liuyer,wangkehang\}@mail.ustc.edu.cn};\\ \normalsize \texttt{\{kkzhang08,qiliuql,cheneh\}@ustc.edu.cn } \\}

\begin{document}
\maketitle
\begin{abstract}
Entity Linking (EL)  is the process of associating ambiguous textual mentions to specific entities in a knowledge base.
Traditional EL methods heavily rely on large datasets to enhance their performance, a dependency that becomes problematic in the context of few-shot entity linking, where only a limited number of examples are available for training. 
To address this challenge, we present OneNet, an innovative framework that utilizes the few-shot learning capabilities of Large Language Models (LLMs) without the need for fine-tuning.
To the best of our knowledge, this marks a pioneering approach to applying LLMs to few-shot entity linking tasks.
OneNet is structured around three key components prompted by LLMs:
(1) an entity reduction processor that simplifies inputs by summarizing and filtering out irrelevant entities, (2) a dual-perspective entity linker that combines contextual cues and prior knowledge for precise entity linking, and (3) an entity consensus judger that employs a unique consistency algorithm to alleviate the hallucination in the entity linking reasoning.
Comprehensive evaluations across seven benchmark datasets reveal that OneNet outperforms current state-of-the-art entity linking methods.

\end{abstract}

\section{Introduction} \label{Intro}
\begin{figure}[!htb]
    \centering
    % \vspace{-1.2cm}
    \includegraphics[width=3.0in]{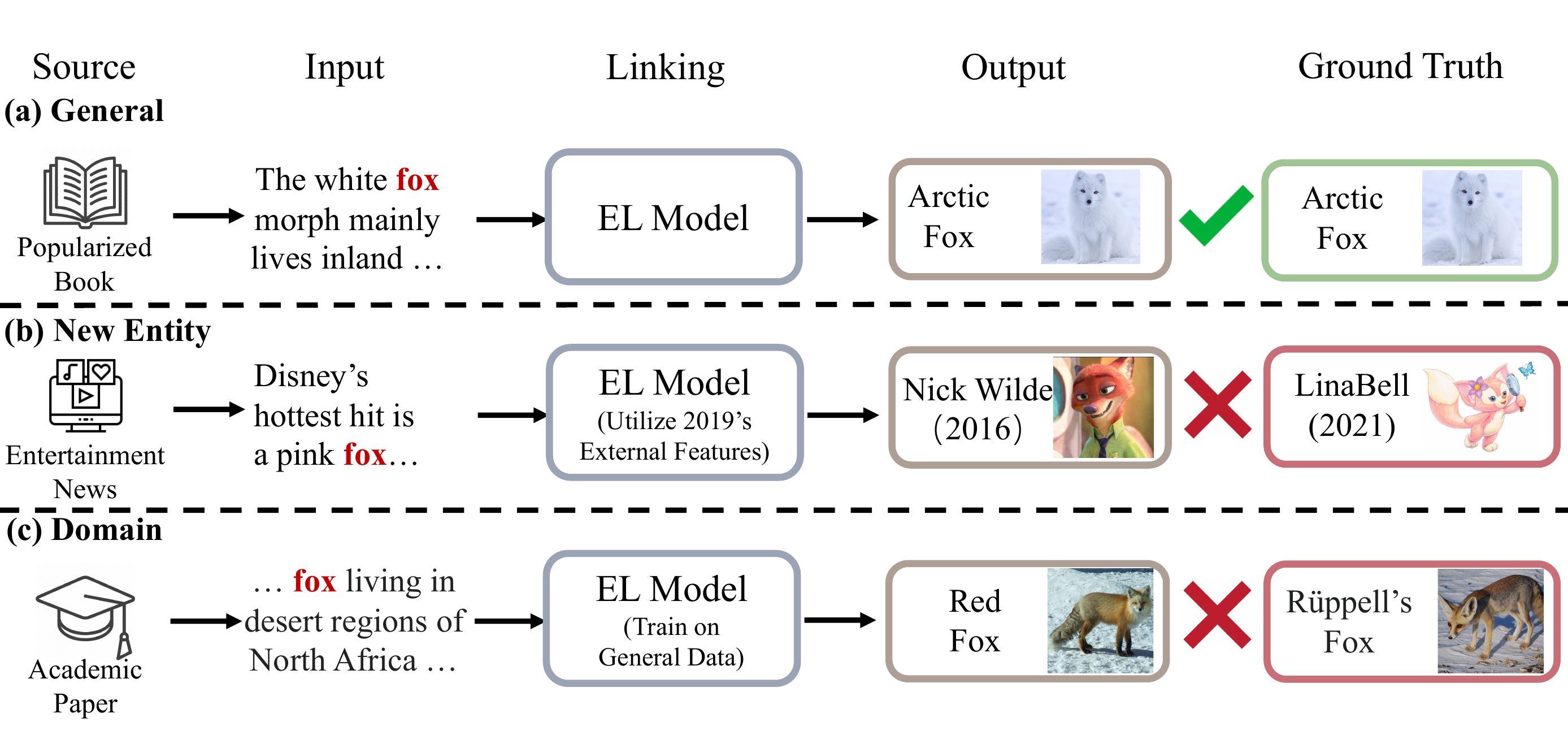}
    \caption{An example of entity linking across various scenarios, where \textbf{mention} is bolded in red.}
    \label{fig:example}
\end{figure}

Entity Linking (EL), also known as Named Entity Disambiguation (NED), entails the process of linking ambiguous textual mentions to specific entities in a knowledge base, as shown in Figure~\ref{fig:example} (a). This process is a critical element of both Natural Language Processing (NLP) and Information Retrieval (IR)~\cite{elsurvey22,liu-etal-2023-enhancing}. 

To enhance the accuracy of EL, researchers employ two primary methods: discriminative models and generative models. 
Discriminative models represent mentions and entities through embeddings and link entities by calculating the similarity. To augment the quality of embeddings, an expansive array of external features are adopted, including categorization of entities~\cite{ner4el21}, imposition of hierarchical constraints~\cite{kehang23}, and incorporation of pre-existing hyperlinks prior~\cite{refined22}.
Conversely, generative models inherently produce the linked entities by adjusting pre-trained language models through fine-tuning processes. These approaches formulate entity linking as various generative tasks, such as sequence-to-sequence constrained generation~\cite{genre21}, information extraction~\cite{extend22}, question answering~\cite{entqa21}, and instruction tuning~\cite{insgen23}, to refine the performance.

Despite promising results shown by conventional approaches, their pronounced reliance on extensive datasets limits their applicability in few-shot scenarios, where only limited annotated examples are available~\cite{xu2023fewshot}. This limitation manifests in two primary ways: Firstly, these methods depend heavily on predefined external features, which compromise their ability to accurately identify novel entities. As depicted in Figure~\ref{fig:example} (b), the reliance on characteristics gathered in 2019 may lead to incorrect linking, such as misidentifying the 2021 Disney fox character, \textit{Linnaeus Bell}, as the character \textit{Nick Wilde} from 2016 due to outdated external features.
Secondly, the dependency on large-scale training datasets poses significant challenges when adapting these models to specialized domains. Figure~\ref{fig:example} (c) demonstrates that limited domain-specific data causes the model to be biased toward general data, potentially causing linking rare species like \textit{Rüppell's fox} to more common ones like \textit{Red fox}.

To mitigate the limitations inherent in conventional models that heavily rely on data, this study explores the utility of Large Language Models (LLMs) to enhance entity linking under a few-shot learning framework. The rationale for employing LLMs is grounded in their significant benefits:
Firstly, LLMs excel at in-context learning~\cite{dong2022surveyICL}, which allows them to reason about previously unencountered questions using a minimal set of examples. 
Second, during pre-training on rich datasets (e.g., Wikipedia), LLMs have acquired versatile prior knowledge, ensuring their proficiency across a multitude of specialized fields~\cite{touvron2023llama}.

However, the utilization of LLMs for entity linking encounters several notable challenges. \textbf{1) Token Length Limitations}. Entity linking typically recalls a substantial volume of candidate entities and their corresponding descriptions, often exceeding 10,000 tokens, which surpasses the capacity of most LLMs. \textbf{2) Reasoning Balance.} Enhanced reasoning techniques such as CoT~\cite{wei2022chainofthought} could inadvertently suppress a model's inherent knowledge~\cite{wei2023largerprioroverwrite}. Maintaining a balance between analytical inference and prior knowledge is crucial for effective EL. \textbf{3) Hallucination.} Hallucination in LLM reasoning emerges as a critical obstacle, especially for complex reasoning tasks such as EL.~\cite{ji2023survey_hallucination}. Identifying and rectifying such reasoning errors in entity linking poses a persistent challenge. 

To tackle these obstacles, we introduce \textbf{OneNet}, a comprehensive framework composed of various interconnected modules, each prompted by LLMs. To our knowledge, in few-shot entity linking, this constitutes the inaugural effort to apply LLMs without fine-tuning. 
Specifically, our approach first begins with the innovative Entity Reduction Processor (ERP), which is designed to condense the input text by summarizing entity descriptions and filtering irrelevant entities.
% Second, to keep the reasoning balanced, we develop the Dual-perspective Entity Linker (DEL). This component undertakes EL by considering both contextual information and prior knowledge.
Second, to maintain an equilibrium in the analytical process, we introduce the Dual-perspective Entity Linker (DEL), which executes EL by integrating contextual cues with prior knowledge.
Third, we address the hallucination problem in EL through our Entity Consensus Judger (ECJ). It undertakes a comparative analysis of two results from DEL, and further employs a consistency algorithm to rectify errors in the reasoning process of LLMs.
Finally, the efficacy of OneNet is underscored across seven diverse datasets, which demonstrate the superiority of our proposed method. 
% We will make our source code publicly available upon the acceptance of our work. 
All prompts are shown in Appendix~\ref{Ap:prompt}, while source codes are available at 
https://github.com/laquabe/OneNet.
\section{Related Works}
\begin{figure*}[t]
    \centering
    \includegraphics[width=\textwidth]{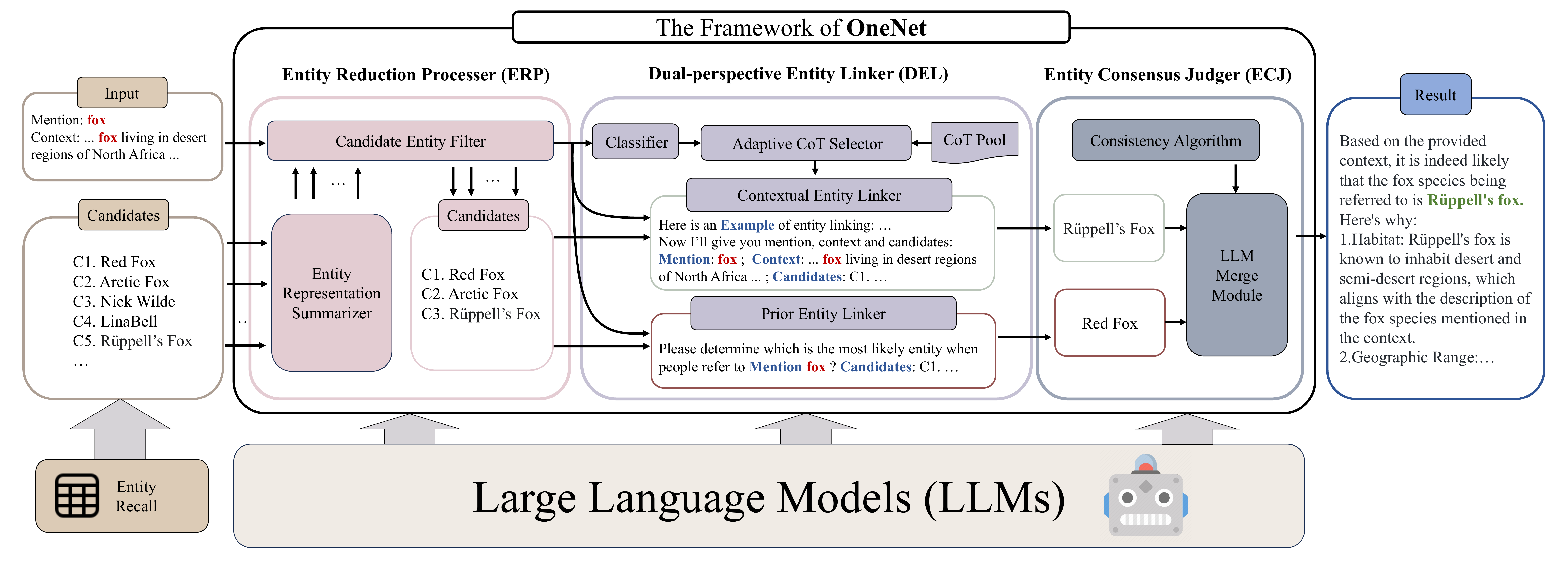}
    \caption{The illustration of OneNet framework, which contains three distinct modules: (a) Entity Reduction Processor (ERP), (b) Dual-perspective Entity Linker (DEL), and (c) Entity Consensus Judger (ECJ).}
    \label{fig:framework}
\end{figure*}
\subsection{Entity Linking} 

Recently, knowledge graphs (KGs) have received widespread attention~\cite{liu-etal-2023-enhancing,liu-etal-2023-rhgn}. Entity linking (EL), which is the core step in constructing KG, has been applied to various fields~\cite{xu2023improving,shi-etal-2024-generative}. As a crucial tool for information extraction and natural language processing~\cite{zhangkai2021eatn,LiuYeTKDD}, early EL studies utilized discriminative models, incorporating external datasets to enhance entity representation. Techniques included hyperlink counts in Deep-ed~\cite{deeped17} and Mulrel-nel~\cite{mulrel18}, NER classifiers in NER4EL~\cite{ner4el21}, and hierarchical constraints in CDHCN~\cite{kehang23}. Other methods leveraged large-scale datasets (e.g., Wikipedia) to boost performance. Blink~\cite{blink20} trained bi-encoders on 5.9 million entities, incorporating titles and descriptions. EntQA~\cite{entqa21} improved this with question-answering techniques, while ReFinED~\cite{refined22} fused priors, types, and descriptions using over 6 million entities. However, these methods depend on extensive data, limiting their ability to novel or domain-specific entities.

Generative models for EL have also emerged recently~\cite{wang-etal-2023-benchmarking}. Genre~\cite{genre21} generates predicted entities after the mention using a constrained decoder. Extend~\cite{extend22} extracts linking entities from context using candidates, while InsGen~\cite{insgen23} applies instruction-tuning on large language models (LLMs). Despite these advances, reliance on fine-tuning still restricts flexibility for few-shot adaptation across diverse scenarios.

Notably, previous studies~\cite{zhou2024gendecider,xu2023read} claimed the suitability of their methods for few-shot and zero-shot learning, yet primarily in out-of-domain contexts. In contrast, our study focuses on a more realistic few-shot framework~\cite{xu2023fewshot}. To the extent of our knowledge, this study is the inaugural exploration of leveraging LLMs for the few-shot entity linking without any necessity for model fine-tuning.

\subsection{Large Language Models}

With the development of pre-training techniques~\cite{zhangkai-etal-2022-incorporating}, large language models (LLMs), including GPT~\cite{achiam2023gpt}, Llama~\cite{touvron2023llama}, and GLM~\cite{du2022glm}, demonstrate impressive few-shot learning ability in numerous natural language processing (NLP) tasks~\cite{liu2023promptfewshotsurvey,zhao-etal-2024-repair}. This emergent capacity allows them to outperform earlier supervised approaches and even achieve human-level performance on certain tasks, all without fine-tuning~\cite{wei2022emergent}.

However, applying LLMs to complex problem-solving remains challenging~\cite{feng2023diffapply}. One way to improve LLM's reasoning is Chain-of-Thought (CoT)~\cite{wei2022chainofthought}, which has attracted growing interest. 
Some research explored optimizing example selection based on similarity~\cite{rubin2022simcot1}, diversity~\cite{zhang2022diversitycot1}, and complexity~\cite{fu2022complexitycot}. 
Other efforts were directed at designing effective reasoning pipelines. For instance, Least2Most~\cite{zhou2022least2most} suggested simplifying complex problems into manageable subproblems. SICoT~\cite{creswell2022selection2inference} proposed a Selection-Inference framework. Furthermore, Deductive CoT~\cite{ling2023dcot} addressed hallucination issues through a sequential reasoning verification process.
Despite these advancements, the application of LLMs in EL necessitates additional investigation.

\section{Preliminary}
\subsection{Few-shot Entity Linking}
In this paper, we formally define $m$ as a mention in a text $\mathcal{S}$, and $e$ as an entity in a knowledge base (KB) associated with its description. For each mention $m$, we have a pre-processing step called entity candidate generation that chooses potential candidate entities $\theta = \{ e_1 , e_2 , ... , e_n \}$ from a specific KB. Each mention $m$ also has a labeled link entity $y$. Following the few-shot setting~\cite{xu2023fewshot}, the training set $D_{train} = \{(S, m, \theta, y )\}$ contains only a few examples and satisfies $|D_{train}| \ll |D_{test}|$. Our goal is to learn the input $(S, m, \theta)$ to output $y$ mapping with as little training data as possible. 

\subsection{Entity Linking with LLMs}

\begin{figure}
    \centering
    \includegraphics[width=3in]{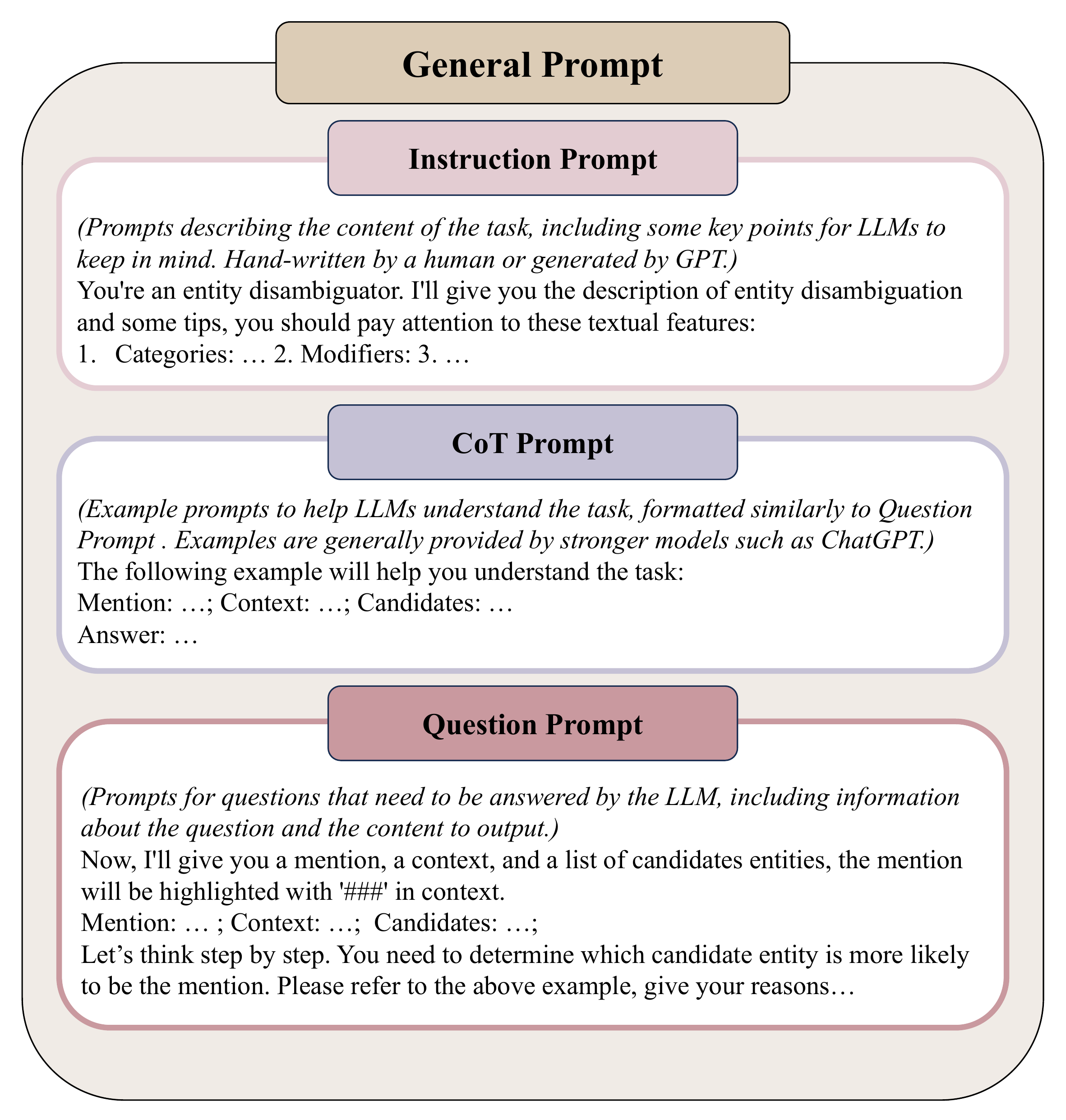}
    \caption{The Illustration of General Prompt Structure}
    \label{fig:prompt_example}
\end{figure}

As illustrated in Figure~\ref{fig:prompt_example}, we form a query for LLMs as $q = [m;S;\theta]$, and the prompt $P$ of entity linking can be composed as a task-specific instruction $I$, $n$ CoT exemplars and the test query itself:

\begin{equation}
    P = [I;q_1;y_1,...,q_n;y_n;q_{test}],
\end{equation}
where $y=(e, r)$ is the output of LLMs, which contains a predicted entity $e$ and the reasoning $r$.

It is important to acknowledge that the difficulties outlined in Section~\ref{Intro} present substantial impediments to the entity linking process when a single Large Language Model (LLM) is employed.
Therefore, we form a pipeline to complete the entity linking by adjusting instruction $I$ and exemplars to prompt multiple modules with different functions. Specific prompts can be found in the Appendix~\ref{Ap:prompt}.

\section{Method}
% In this section, we first present the problem of entity linking with LLMs, followed by an overview of our framework. Then we introduce the technical details of our framework.

\subsection{An Overview of One-Net}

As shown in Figure~\ref{fig:framework}, the proposed methodology comprises three distinct modules: (a) Entity Reduction Processor (ERP), (b) Dual-perspective Entity Linker (DEL), and (c) Entity Consensus Judger (ECJ). 
Initially, ERP conducts a two-step process that involves the summarizing of entity descriptions and the point-wise exclusion of irrelevant candidate entities. 
Subsequently, DEL is devised to establish fine-grained entity linking within the filtered candidates, utilizing both contextual analysis and prior knowledge. 
Finally, ECJ combines the contextual result and prior result to generate the predicted entity.
Notably, each module is derived from a large language model, leveraging distinct prompts without fine-tuning.

\subsection{Entity Reduction Processor} \label{ERP}
To address the issue of token length limitations in Section~\ref{Intro}, we employed the following optimization strategy. As illustrated in Figure~\ref{fig:framework}, firstly, the Entity Representation Aggregator is utilized to condense the descriptions of entities, thereby providing a more succinct representation. Secondly, the Candidate Entity Filter is implemented to execute an initial, point-wise filtration of potential entities to reduce the number of candidates. 

\subsubsection{Entity Representation Summarizer}
From previous research~\cite{cheng2015summarizing}, entity summarization can significantly enhance the efficiency of entity linking by distilling the essential characteristics of entity descriptions. In this context, we function as a summarizer by engaging a large language model through a simple prompt. The prompt $P_{sum}$ consists solely of a summary instruction prompt and an entity, which are structured as follows:
\begin{equation}
    P_{sum} = [I_{sum};e],
\end{equation}
where $I_{sum}$ is the summary instruction prompt, $e$ is an entity with its description.

\subsubsection{Candidate Entity Filter} \label{Sec: Filter}
In light of the suboptimal performance exhibited by directly list-wise EL, the Candidate Entity Filter transfers the list-wise EL into a sequence of point-wise EL, which only has one candidate in the query. This strategic conversion facilitates the effective filtration of irrelevant entities, which prompt $P_{fil}$ is as follows:
\begin{equation}
    P_{fil} = [I_{el};m;S;e_i],
\end{equation}
where $I_{el}$ is the instruction of entity linking, $e_i$ is one entity in the candidates. To improve efficiency, we don't use the Chain-of-Thought methods.
Inspired by the insights of prior research~\cite{honovich2022instruction}, we utilize LLMs to formulate instructions, which details are shown in Appendix~\ref{Ap:prompt}.

\subsection{Dual-perspective Entity Linker} \label{DEL}

Building upon the established understanding from prior research~\cite{deeped17, mulrel18}, it is recognized that entity linking can be decomposed into two distinct components: a prior probability $p(e)$ and a contextual probability $p(c|e)$.
Accordingly, to maintain an equilibrium in the analytical process mentioned in Section~\ref{Intro}, the Dual-perspective Entity Linker is composed of two components in Figure~\ref{fig:framework}: the Contextual Entity Linker, which leverages the inference capabilities of LLMs to generate context-aware predictions, and the Prior Entity Linker, which employs the inherent knowledge embedded within LLMs to produce predictions based on prior information.
% which can be succinctly expressed through the following formalization:
% \begin{equation}
%     P(e | c) \propto P(e)P(c | e),
% \end{equation}
% where $P(e)$ is the link probability without context, $P(c|e)$ can be seen as the relevance of the entity and the context.

\begin{algorithm}
	\renewcommand{\algorithmicrequire}{\textbf{Input:}}
	\renewcommand{\algorithmicensure}{\textbf{Output:}}
	\caption{The Consistency Algorithm}
	\label{alg:1}
	\begin{algorithmic}[1]
		\REQUIRE contextual prediction $e_{context}$, prior prediction $e_{prior}$, mention $m$, context $S$
		\ENSURE link entity $e_{result}$
		\IF{$e_{context} = e_{prior}$}
                \STATE $e_{result} \gets e_{context}$
            \ELSE
                \STATE $e_{result}\gets$ ${LLM}(e_{context}, e_{prior}, m, S)$
            \ENDIF
	\STATE \textbf{return} $e_{result}$
	\end{algorithmic}  
\end{algorithm}

\subsubsection{Contextual Entity Linker}
To effectively harness the inferential capabilities of LLMs for list-wise entity linking, the Contextual Entity Linker employs a structured prompt as depicted in Figure~\ref{fig:prompt_example}. This prompt is composed of three distinct segments: the instruction prompt, the CoT prompts, and the question prompt, which is formed as:
\begin{equation}
    P_{context} = [I_{el};q_1;y_1,...,q_n;y_n;q_{test}],
\end{equation}
where $I_{el}$ is the entity linking instruction mentioned in Section~\ref{Sec: Filter}, $[q_i,y_i]$ is the CoT exemplar.

\noindent \textbf{CoT Exemplar Pool.} Inspired by previous work~\cite{webglm}, in order to distill the reasoning power of the advanced models, we sample a subset of questions from the training dataset and present them to advanced models for response. To mitigate the issue of hallucination\cite{ji2023survey_hallucination}, we implement a stringent selection criterion, retaining only those responses that accurately predict the correct entity.

\noindent \textbf{Adaptive CoT Selector.} To effectively determine the optimal CoT reasoning approach, our selection process is informed by two critical dimensions: context similarity and entity category.
Firstly, we postulate that similar contexts likely share analogous reasoning patterns. To implement this, we quantify the resemblance by computing the cosine similarity between the input context and the exemplar contexts in our CoT pool.
Secondly, we recognize that mentions belonging to the same category often exhibit common features that are pertinent to the reasoning process. To leverage this, we employ an LLM as a classifier, which incorporates a specific classifier instruction prompt along with the mention and its provided context.
Ultimately, our composite CoT score is derived by integrating these considerations:
\begin{equation}
    s = \alpha \cdot cos(S_{i}, S_{test}) + (1-\alpha) \cdot \mathbb I(m_i, m_{test}),
\end{equation}
where $\mathbb I(\cdot,\cdot)$ indicates whether the category is the same in both mentions and $\alpha$ is a hyperparameter.
\subsubsection{Prior Entity Linker}
To utilize the inherent prior knowledge in the LLMs, we employ an LLM as the Prior Entity Linker. As shown in figure~\ref{fig:framework}, the prior prompt is comprised of three distinct components: the prior instruction prompt, the mention, and the filtered candidates, which can be represented as follows:
\begin{equation}
    P_{prior} = [I_{prior};m;\theta_{fil}].
\end{equation}

It is worth noting that the context is hidden to prevent the influence of noise in the context on the prior~\cite{noisy_EL}. Due to the lack of context, many of the hints about the context in the entity linking instructions are no longer appropriate, so we use instructions that are not the same as the contextual linker. Additionally, the imperative for preserving prior knowledge necessitates the exclusion of CoT methods to preclude the potential overwriting of LLM intrinsic knowledge~\cite{wei2023largerprioroverwrite}.

\subsection{Entity Consensus Judger}
To ensure accurate entity prediction from the two predicted entities in Section~\ref{DEL}, the Entity Consensus Judger utilizes a consistency algorithm to mitigate potential hallucination in DEL, as illustrated in Figure~\ref{fig:framework}. The algorithm functions as follows: when both prediction modules concur on the same entity, that entity is confirmed as the result. Conversely, in instances of prediction discordance, the ECJ invokes an auxiliary LLM to ascertain the correct entity for linking. The details of this algorithm are shown in Algorithm~\ref{alg:1}.

The propensity for inaccuracies within the Contextual Linker predominantly stems from misleading of CoT. Conversely, errors within the Prior Linker are principally attributed to the lack of context. To mitigate the occurrence of both error types, the auxiliary LLM has been designed to incorporate instruction prompt, context, and the entities ascertained by the dual linkers, which is formed as:
\begin{equation}
    P_{merge} = [I_{el}; m; S; e_{context}; e_{prior}],
\end{equation}
where $I_{el}$ is the entity linking prompts as Section~\ref{Sec: Filter}. Candidate entities are limited to the entities predicted by the previous linkers.

\begin{table}[ht]
\centering
\resizebox{\linewidth}{!}{
\begin{tabular}{c|c c c c c} \toprule
\textbf{Dateset} & \textbf{Mentions} & \textbf{Cand. Nums} & \textbf{Ent. Tokens} & \textbf{Cont. Tokens} & \textbf{Alias Recall} \\ \midrule
\textbf{ACE2004} & 257 & 42.25 & 190.79 & 171.17 & 0.977 \\
\textbf{AIDA}	& 4463 & 7.18 & 262.16 & 452.53 & 0.982 \\
\textbf{AQUAINT} &  727 & 34.69 & 197.00 & 169.57 & 0.905 \\
\textbf{CWEB} &  11154 & 6.98 & 239.36 & 222.126 & 0.948 \\ 
\textbf{MSNBC} & 656 & 25.93 & 198.81 & 198.88 & 0.982 \\
\textbf{WIKI} & 6821 & 6.09 & 227.04 & 195.90 & 0.956 \\
\textbf{ZeShEL} & 10000 & 55.20 & 441.65 & 2394.93 & 0.681 \\
\bottomrule
\end{tabular}
}
\caption{The Statistics of Test Datasets}
\label{table:statistics}
\end{table}

\section{Experiments}

\subsection{Datasets}
For the reliability and authority of experimental results, we have conducted evaluations across seven widely recognized datasets: ACE2004, AIDA\cite{aida}, AQUAINT\cite{guo2018robustace2004}, CWEB, MSNBC, WIKI~\cite{cweb}, and ZeShEL~\cite{logeswaran-etal-2019-zeshel}. Table~\ref{table:statistics} provides further information about the datasets. As our method utilizes LLMs, we also calculated the tokens to estimate costs. For Wiki-based datasets, We utilize the November 2020 snapshot of English Wikipedia~\cite{ner4el21} as our knowledge base (KB). Following the previous work~\cite{kehang23}, we employed an alias table to generate the candidate entities. For efficiency, we limited the candidate pool to 10 entities for datasets with large numbers of mentions, such as AIDA, CWEB, and WIKI. For the remaining Wiki-based datasets, we retained the complete set of candidate entities. For ZeShEL, we used the top-64 TFIDF candidates provided by the authors. We also measured alias table recall to assess quality.

\begin{table*}[!ht]
\centering
\resizebox{13.5cm}{!}{
\begin{threeparttable}
\renewcommand\arraystretch{0.9}
\begin{tabular}{c c | c c c c c c } \toprule
\multicolumn{2}{c|}{\textbf{Dateset}} & \textbf{ACE2004} & \textbf{AIDA} & \textbf{AQUAINT} & \textbf{CWEB} & \textbf{MSNBC} & \textbf{WIKI} \\ \midrule
\multirow{4}{*}{Tradition}&Mulrel-nel & 0.217 & 0.328 & 0.262 & 0.267 & 0.422 & 0.380 \\
&NER4EL & 0.531 & 0.569 & 0.460 & 0.488 & 0.602 & 0.495 \\ 
&Extend & 0.604 &0.563 & 0.641 &0.537 &0.715 & 0.506 \\ 
&CDHCN & 0.438 & 0.575 &0.465 &0.504 & 0.654 & 0.505 \\ \midrule
\multirow{4}{*}{LLMs}&GLM-8K\tnote{1} & 0.482 & 0.520 & 0.466 & 0.454 & 0.550 & 0.550 \\ 
&GLM-32K & 0.447 & 0.439 & 0.431 & 0.487 & 0.584 & 0.532 \\ 
&Zephyr & 0.467 & 0.322 & 0.495 & 0.518 & 0.637 & 0.555 \\ 
&ChatGPT & 0.611 & 0.451 & 0.560 & 0.546 & 0.732 & 0.615 \\ \midrule
\multirow{3}{*}{OneNet} &GLM-8K &0.611	& 0.639	& 0.626	& 0.587	& 0.713	& 0.626 \\ 
&GLM-32K &0.650	& 0.672	& 0.626	& 0.606	& 0.764	& 0.645 \\ 
&Zephyr & \textbf{0.681} & \textbf{0.690} & \textbf{0.686} & \textbf{0.650} & 
\textbf{0.796} & \textbf{0.676} \\ 
\bottomrule
\end{tabular}
\begin{tablenotes}
\footnotesize
       \item[1] The data used for GLM-8K is filtered by our Entity Reduction Processor.
\end{tablenotes}
\end{threeparttable}}
\caption{Micro-F1 Scores of Few-shot Entity Linking on Various Datasets}
\label{table:main}
\end{table*}

\begin{table}[!ht]
\centering
\resizebox{\linewidth}{!}{
\begin{tabular}{c|c c c c c c} \toprule
\textbf{Dateset} & \textbf{ACE2004} & \textbf{AIDA} & \textbf{AQUAINT} & \textbf{CWEB} & \textbf{MSNBC} & \textbf{WIKI} \\ \midrule
ERP Recall & 0.765 & 0.885 & 0.844 &0.802 & 0.880 & 0.875 \\
Filtering Rate & 0.900 & 0.648 & 0.865 & 0.643 & 0.825 & 0.622 \\
Avg & 4.07 & 2.55 & 4.72 & 2.51 & 4.57 & 2.31 \\ \bottomrule
\end{tabular}}
\caption{Recall, Filtering Rate, and Average Candidates across Datasets Filtered by ERP.}
\label{table:recall}
\end{table}

\subsection{Implementation Details} 
For Wiki-based datasets, we implement our method on Zephyr-7b-beta~\cite{tunstall2023zephyr} and GLM~\cite{du2022glm}. The exemplar pool, comprising 65 data instances, is derived from the training set of AIDA. We place $n = 1$ exemplar in the prompt $P$ for the contextual entity linker. The adaptive CoT selector's hyper-parameter is set to $\alpha = 0.5$. When running Zephyr, we fix the parameters to the default values provided by the official, 
% where the sampling temperature is 0.7, top\_k is 50, and top\_p is 0.95. 
and the max new token is set to 1024.
For classifier, we use Wikipedia's 12 categories.
To mitigate the potential bias arising from sequence dependency within the model, we randomize the order of candidate entities for each time. We take the first occurrence of the entity as the prediction.
Following the previous work~\cite{elsurvey22}, we report the micro F1 to assess entity linking performance. For ZeShEL, we use the same setting as Wiki-based datasets, which is described in appendix~\ref{Ap:general}.

\subsection{Benchmark Methods} \label{benchmark}
To evaluate the effectiveness of One-Net, we compare it with traditional state-of-the-art supervised methods and popular large language models:
\begin{itemize}
\item \textbf{Traditional Supervised Methods.} These models necessitate supervised learning. Specifically, Mulrel-nel~\cite{mulrel18}, NER4EL~\cite{ner4el21}, and CDHCN~\cite{kehang23} utilize leverage external data to train discriminative models. Conversely, Extend~\cite{extend22} employs a generative approach to extract the corresponding entity from the candidate entities. Our exemplar pool provides the foundational data required for the training.

\item \textbf{Large Language Models.} Since entity linking is a text-only task, LLMs can also be directly applied to it. For GLM~\cite{du2022glm}, We tested both 4K and 32K versions to confirm the effectiveness of long text training. For Zephyr~\cite{tunstall2023zephyr}, we use the same version to validate the effectiveness of our framework. For ChatGPT~\cite{GPT1}, we utilize the model \textit{gpt-3.5-turbo-1106} to test. During the generation of outputs, we adhered to the default settings provided by the official documentation. The same exemplars are provided to all LLMs to facilitate their chain-of-thought ability.

\end{itemize}

As we focus on the few-shot scenario, we disregard additional models which are trained on massive additional data, such as Blink~\cite{blink20}, EntQA~\cite{entqa21}, and ReFinED~\cite{refined22}, to ensure an equitable comparison. Additionally, we have omitted results from other popular language models like Llama2~\cite{touvron2023llama}, as their performance is found to be suboptimal, which falls below 5\%.

\subsection{Experimental Results}
The results of all methods on the datasets are shown in Table \ref{table:main} and Table~\ref{table:ZeShEL}. For Wiki-based datasets, We report three OneNet results based on different base models. In general, OneNet with Zephyr has achieved the best performance compared with SOTA baselines. For ZeShEL, OneNet also achieves optimal results on most domains. Specifically, our method outperforms the best-performing baseline (i.e., Extend, ChatGPT) by 4\%-11\%. Additionally, we discover some interesting phenomena:

First, traditional generative models, such as Extend, demonstrate superior performance over traditional discriminative models like NER4EL and CDHCN in few-shot scenarios, supporting the idea of generative LLMs for few-shot entity linking. 
Second, a single LLM does not have good entity linking capability. For instance, all of OneNet's results are significantly better than the corresponding single model's results (i.e., GLM, Zephyr). 
Third, when switching to ZeShEL, the performance of a single LLM is worse due to the excessive length of inputs and the numerous irrelevant entities, which negatively impact the model. In contrast, OneNet demonstrates robust performance, outperforming traditional methods.
These findings underscore the necessity of using multiple prompts to leverage the diverse capabilities of LLMs for effective EL. 

\subsection{Ablation Study}

\subsubsection{Accuracy and Efficiency of ERP}

To validate the effectiveness of the ERP in Section~\ref{ERP}, we show the recall, filtering rate, and average number of remaining candidate entities on all datasets, where filter rating shows the percentage of filtered-out candidates. In fact, as the first module of OneNet, ERP determines the performance ceiling of the entire pipeline. As shown in Table~\ref{table:recall}, recall reaches 0.8 for most datasets except ACE2004, and for filtering rate and Avg, the filtering rate reaches more than 0.8 on the unprocessed dataset, and the Avg is around 4. For the preprocessed dataset, the filtering rate reaches more than 0.6 and the Avg is around 3. All these prove that ERP can filter out as many irrelevant entities as possible while ensuring that the correct entities are retained.

\begin{figure}
    \centering
    \includegraphics[width=2.7in]{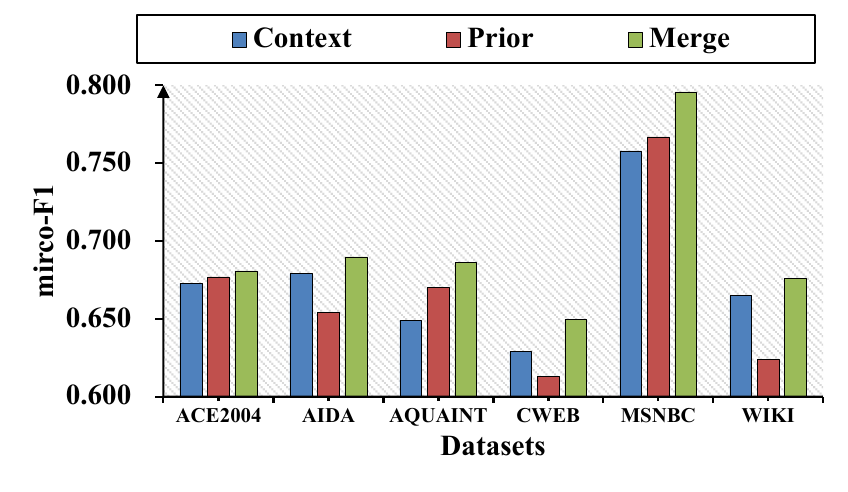}
    \caption{Comparison of Prior, Context, and Merge}
    \label{fig:merge}
\end{figure}

\begin{figure*}[!ht]
    \centering
    \vspace{-0.5cm}
    \includegraphics[width=\textwidth]{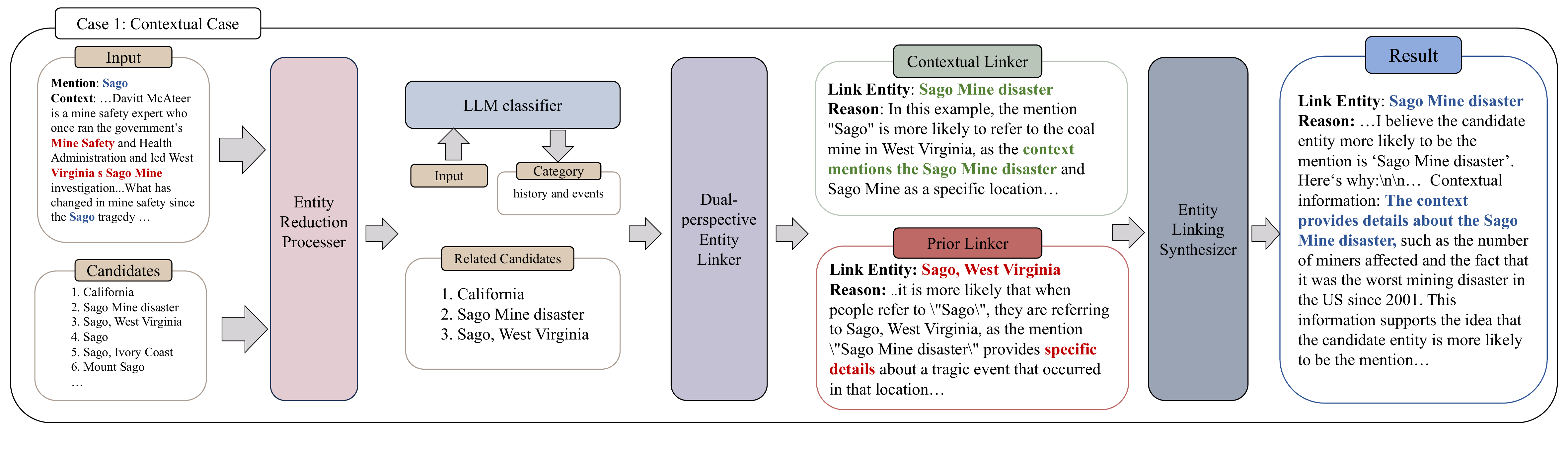}
    \caption{Case study of OneNet. Key information in context makes the contextual linker reason correctly.}
    \label{fig:case_study}
\end{figure*}

\begin{figure}
    \centering
    \includegraphics[width=2.7in]{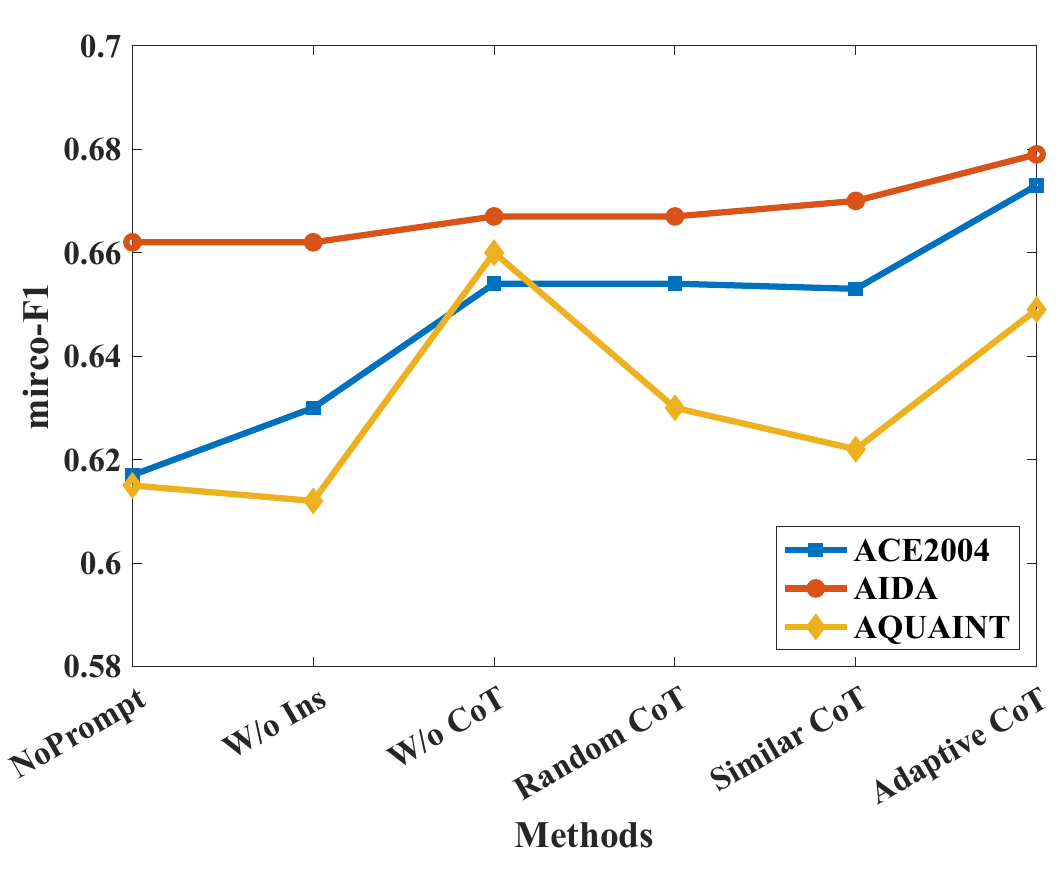}
    \caption{The Impact of Different Prompt Structure}
    \label{fig:COT_method}
\end{figure}

\subsubsection{Context and Prior are Both Necessary}
As we mentioned in Section~\ref{DEL}, to substantiate the indispensability of both context and prior perspectives, a comparative analysis of the individual modules and their merged results is conducted. Figure~\ref{fig:merge} illustrates that merge results yield superior performance across all datasets, thereby affirming ECJ's efficacy. Nonetheless, it is noteworthy that the context and prior components each exhibit distinct advantages across varying datasets. For example, context outperforms 2\% on AIDA, while prior is 2\% higher on AQUAINT. This observation validates the rationale behind incorporating dual perspectives within the DEL module.

\subsubsection{Detailed Instructions and Reasonable Exemplars make LLMs Aligned} \label{Sec:prompt_exp}

Figure~\ref{fig:prompt_example} elucidates that our prompt contains both detailed instructions and reasonable exemplars to facilitate LLM's understanding of entity linking.
The comparative results, as summarized in Figure~\ref{fig:COT_method}, demonstrate that our adaptive CoT approach surpasses other CoT selection methods across all evaluated datasets, which underscores the efficacy of our method in identifying more suitable exemplars.
Meanwhile, our findings indicate that the absence of detailed instructions hampers the LLM's ability to understand the EL task (e.g., NoPrompt, W/o Ins).
Furthermore, our analysis reveals that prompts with CoT demonstrate superior performance in ACE2004 and AIDA. Conversely, prompts without CoT exhibit enhanced efficacy in AQUAINT. This result is consistent with Figure~\ref{fig:merge}, as AIDA is more context-aware, AQUAINT is more prior-dependent and ACE2004 considers both. This further suggests that both reasoning ability and prior knowledge are important for EL.

To test the robustness of our module, we generate various instruction prompts for testing. The results, illustrated in Figure~\ref{fig:diff_prompt} in Appendix~\ref{Ap:exp}, 
demonstrate that the module’s performance remained stable despite variations in the prompts. 
Additionally, Table~\ref{table:repeated_answers} in Appendix~\ref{Ap:exp} presents an example of repeated answers, emphasizing that the semantic of outputs is invariant within the framework constraints.
These consistency underscores the user-friendliness of our module, demonstrating its ability to perform reliably under diverse instructions.

\subsection{Case Study of OneNet} \label{Sec:case_study}
For a more intuitive comparison of how our frameworks work, we provide two case studies, one utilizing contextual linking in Figure~\ref{fig:case_study} while the other utilizing prior linking in Figure~\ref{fig:prior_case} in Appendix~\ref{Sec:prior_case}.  

The first mention is \textit{Sago}, found in an article on mining safety.
Initially, the Entity Reduction Processor screened out 6 irrelevant entities. For instance, \textit{Sago} as a foodstuff and \textit{Mount Soga} for its geographical inaccuracy. Subsequently, three pertinent entities remained: \textit{Sago Mine disaster}, \textit{Sago, West Virginia}, and \textit{California}. 
Following the classification \textit{history and events}, the contextual linker identified \textit{Sago Mine disaster} as the likely reference, deducing that \textit{Sago Mine} was implied within the text. Conversely, the prior linker suggested \textit{Sago, West Virginia}, which considers \textit{Sago Mine disaster} to be overly specific.
Ultimately, the Entity Consensus Judger favored the contextual prediction \textit{Sago Mine disaster}, corroborated by the text's detailed description of the event. This resolved an error in the prior linker by taking into account the context provided.

The second entity mentioned is \textit{Orange County} in an airport blog. The procedure mirrors that of the initial case. However, the term "airport" in the context notably causes the contextual linker's error. In contrast, the prior linker predominantly depends on the model's intrinsic knowledge to render an accurate prediction. Details are provided in Figure~\ref{fig:prior_case} in Appendix~\ref{Sec:prior_case}.
% The second entity mentioned is \textit{Orange County}, which appears in an airport blog. Initially, the Entity Reduction Processor filtered out 23 irrelevant entities, such as \textit{Orange County, Florida}, \textit{Orange County (film)}. After this process, three relevant entities remained: \textit{Orange County}, \textit{California}, \textit{John Wayne Airport}.
% Under the classification of \textit{geography and places}, the contextual linker pinpointed \textit{John Wayne Airport} as the probable reference, which thinks that the article's focus is on airports and travel. In contrast, the prior linker posited \textit{Orange County, California} as the more frequent referent in general discourse.
% Ultimately, the Entity Consensus Judger give precedence to the prior linker's prediction of \textit{Orange County, California}. It noted that \textit{John Wayne Airport} is situated within \textit{Orange County, California}, which clarified the confusion for the contextual linker.

More experimental analyses, such as Framework Generalization, can be found in Appendix~\ref{Ap:general}.
\section{Conclusion}
In this study, we introduced OneNet, a novel framework for few-shot entity linking by leveraging large language model prompts without fine-tuning. Specifically, OneNet was comprised of three key LLM-prompted components: 
the Entity Reduction Processor, which was designed for efficient text condensation by summarizing entity descriptions and irrelevant entity filtering;
the Dual-perspective Entity Linker, which considered both contextual information and prior knowledge to provide a balanced analysis; 
and the Entity Consensus Judger, which was instrumental in reducing hallucinations through a consistency merger algorithm.
Our framework demonstrated superior performance on seven datasets. 
Our future research will aim to merge mention detection within our model.

\section{Limitations}
Although we have demonstrated the superiority of our OneNet compared to previous work on
seven real-world datasets, there are still two limitations that should be addressed in the future:

(1) Our framework relies on prompting LLMs, thus its efficiency is constrained by LLM inference speed. As shown in Table~\ref{table:runtime} in Appendix~\ref{Ap:exp}, the runtime is heavily influenced by the base model. 
Table~\ref{table:input_tokens} Appendix~\ref{Ap:exp} reports the average input tokens per module, showing that our framework does not substantially increase token requests compared to direct LLM use. 
Additionally, some modules (e.g., ERS) can run offline, which will enhance efficiency.
Nonetheless, the field has witnessed significant advancements aimed at expediting the inference process for LLMs. These enhancements encompass strategies like I/O optimization~\cite{dao2022flashattention}, model pruning~\cite{liu2023dejaPruning1}, and quantization techniques~\cite{dettmers2022Quant1},. It is our assertion that these ongoing research efforts will eventually surmount the current limitations imposed by the inference speed of large language models, thereby mitigating this bottleneck in the foreseeable future.

(2) Currently, our framework is dedicated exclusively to the task of entity disambiguation. It is important to note that the broader domain of entity linking encompasses both entity disambiguation and mention detection. Actually, mention detection has been effectively approached using large language models~\cite{jin-etal-2023-adversarialllmner1} and prompting techniques~\cite{shen-etal-2023-promptnerllmner2}, its integration is not only complementary but can also enhance the performance of entity disambiguation. In future work, we will explore more efficient ways to integrate entity disambiguation and mention detection.

\section*{Acknowledgements}
This research was supported by grants from the Joint Research Project of the Science and Technology Innovation Community in Yangtze River Delta (No. 2023CSJZN0200), the National Natural Science Foundation of China (No. 62337001), Anhui Provincial Natural Science Foundation (No. 2308085QF229), and the Fundamental Research Funds for the Central Universities.
%(No. WK2150110034)

\bibliography{anthology,custom}
\bibliographystyle{acl_natbib}

\appendix

\section{Experimental Supplement} \label{Ap:exp}

\subsection{Different Instruction Prompts} 

\begin{figure}[ht]
    \centering
    \includegraphics[width=2.7in]{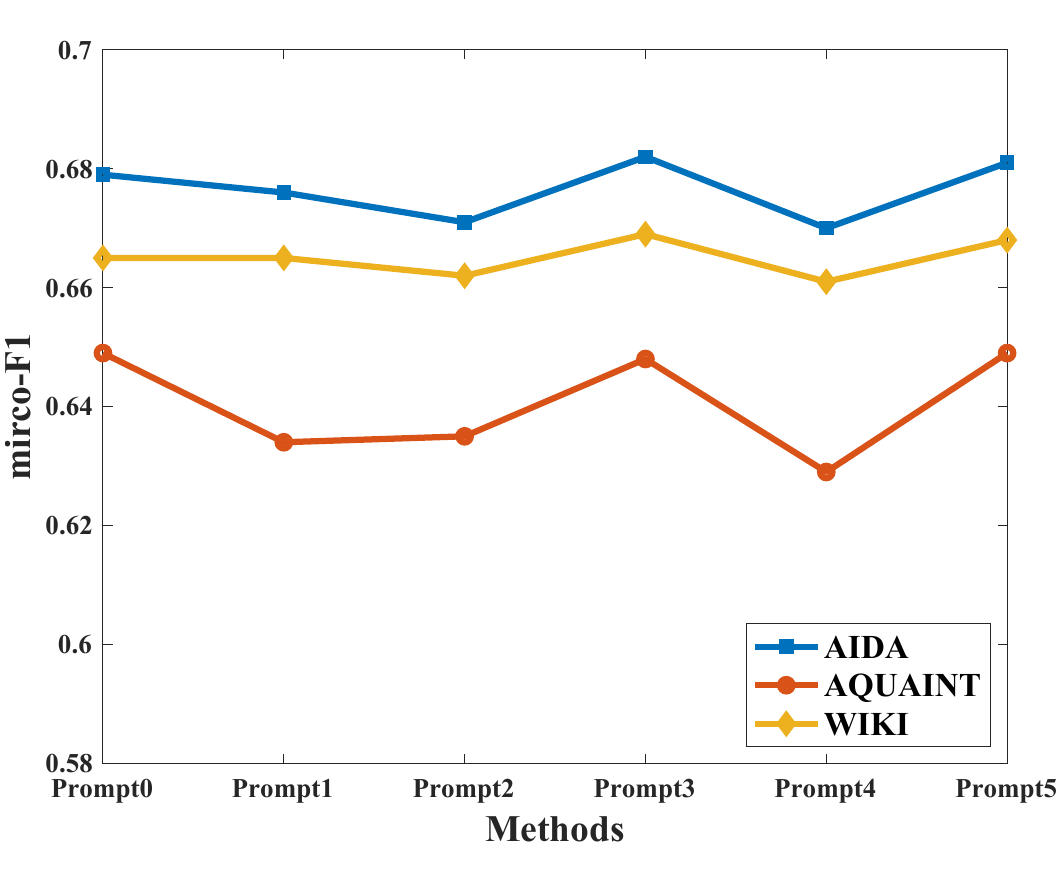}
    \caption{The Impact of Different Instruction Prompts}
    \label{fig:diff_prompt}
\end{figure}

Due to space constraints, Figure~\ref{fig:diff_prompt} mentioned in the main text have been moved to the appendix, which shows the performance of the contextual linker with different instruction prompts in Section~\ref{Sec:prompt_exp}

\subsection{Repeated Answers} 
To provide a more intuitive illustration of the robustness of our framework, we provide a case study of repeated responses. As shown in Table~\ref{table:repeated_answers}, although the expressions of the model outputs are different, none of the semantics of the results change, which demonstrates the stability of our framework.

\begin{table}[!ht]
\centering
\resizebox{\linewidth}{!}{
\begin{tabular}{c|c c c c c c} \toprule
\textbf{Dateset} & \textbf{ACE2004} & \textbf{AIDA} & \textbf{AQUAINT} & \textbf{CWEB} & \textbf{MSNBC} & \textbf{WIKI} \\ \midrule
GLM-8K	& 1.78	& 0.97	& 1.27	& 0.92	& 1.19	& 0.97 \\
GLM-32K	& 1.88	& 1.13	& 1.58	& 1.12	& 1.31	& 1.24 \\
Zephyr	& 15.22	& 11.19	& 15.69	& 13.08	& 14.75	& 14.26 \\ \bottomrule
\end{tabular}}
\caption{Execution Time (s) per Mention with Various Base Models}
\label{table:runtime}
\end{table}

\begin{table}[!ht]
\centering
\resizebox{\linewidth}{!}{
\begin{tabular}{c|c c c c c c c} \toprule
\textbf{Module} & \textbf{ACE2004} & \textbf{AIDA} & \textbf{AQUAINT} & \textbf{CWEB} & \textbf{MSNBC} & \textbf{WIKI} & \textbf{ZeShEL}\\ \midrule
CEF	& 507.79	& 881.56 &	491.06	& 604.36 & 	516.35 & 552.32	& 1073.84 \\
Classifier & 281.88	& 605.46  &	304.78 & 363.35 &	343.94 & 	304.42 &	216.10 \\
CEL	& 1765.35	& 1957.43	& 1734.89	& 1661.61	& 1676.05	& 1662.00	& 3086.82 \\
PEL	& 426.23	& 296.64	& 374.24	& 238.60	& 276.09	& 301.05	& 1841.94 \\
ECJ	& 86.21	& 133.87	& 93.35	& 268.16	& 64.63	& 186.07	& 523.00   \\ \midrule
Raw	& 10268.17	& 4214.20	& 6715.72	& 4493.97	& 5182.58	& 4115.79	& 36470.82 \\
\bottomrule
\end{tabular}}
\caption{Input Tokens for Each Module on Various Datasets}
\label{table:input_tokens}
\end{table}

\begin{figure*}[!ht]
    \centering
    \includegraphics[width=\textwidth]{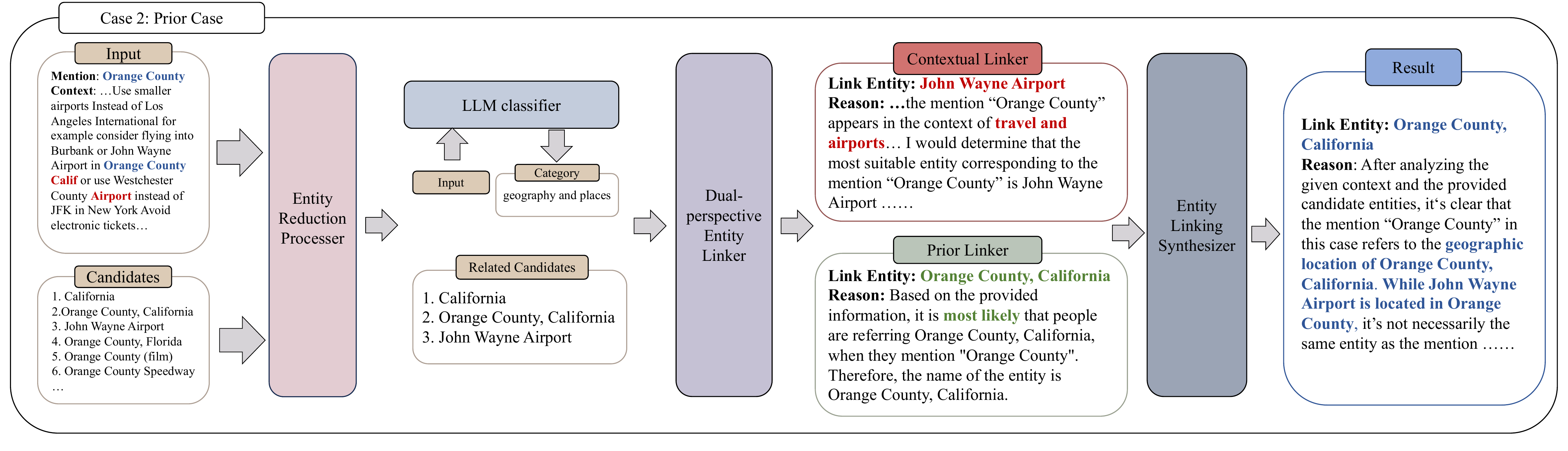}
    \caption{Case Study of OneNet. Prior with LLMs makes the prior linker reason correctly.}
    \label{fig:prior_case}
\end{figure*}

\begin{table*}[htbp] 
\vspace{-0.25cm}
\begin{tabular}{c p{13cm}c} \hline
Mention & \textbf{Tallahassee} \\ \hline
Context & \textbf{Tallahassee} United States 11 15 AFP The Supreme Court in Florida today Wednesday refused the application by the state s authorities to ... \\ \hline
Entity & \textbf{Tallahassee, Florida} \\ \hline
Answer 1 & Based on the given context and the features to consider for entity disambiguation, the best match for the mention Tallahassee appears to be entity \textbf{Tallahassee, Florida}. Here are the reasons for the selection:... \\ \hline
Answer 2 & Based on the context provided and considering the features mentioned, the most suitable entity for the mention Tallahassee is entity \textbf{Tallahassee, Florida}. The context describes Tallahassee as the location where the Supreme Court in Florida refused to stop the new hand count of votes... \\ \hline
Answer 3 & In this case, both candidate entities (Tallahassee, Florida and Tallahassee metropolitan area) are locations, but the mention Tallahassee in the given context is more likely referring to the capital city of Florida rather than the metropolitan area ... Therefore, the serial number for the correct entity is \textbf{"Tallahassee, Florida"} \\ \hline
Answer 4 & Based on the context provided, it is most likely that the mention Tallahassee refers to entity \textbf{Tallahassee, Florida}. The context mentions the Supreme Court in Florida refusing an application to stop a vote count in some counties ... \\ \hline
Answer 5 & Based on the context provided and considering the features mentioned, the most suitable entity for the mention Tallahassee is entity: \textbf{Tallahassee, Florida}. The context states that the Supreme Court in Florida ... \\ \hline
\end{tabular}
\caption{Case Study of Repeated Answers}
\label{table:repeated_answers}
\end{table*}

\subsection{Case Study for Prior} \label{Sec:prior_case}

As we mention in Section~\ref{Sec:case_study}, we also provide another case for prior linking in Figure~\ref{fig:prior_case}. The entity mentioned is \textit{Orange County}, which appears in an airport blog. Initially, the Entity Reduction Processor filtered out 23 irrelevant entities, such as \textit{Orange County, Florida}, \textit{Orange County (film)}. After this process, three relevant entities remained: \textit{Orange County}, \textit{California}, \textit{John Wayne Airport}.
Under the classification of \textit{geography and places}, the contextual linker pinpointed \textit{John Wayne Airport} as the probable reference, which thinks that the article's focus is on airports and travel. In contrast, the prior linker posited \textit{Orange County, California} as the more frequent referent in general discourse.
Ultimately, the Entity Consensus Judger gives precedence to the prior linker's prediction of \textit{Orange County, California}. It noted that \textit{John Wayne Airport} is situated within \textit{Orange County, California}, which clarified the confusion for the contextual linker.

\subsection{Framework Efficiency} 
As mentioned in Limitation, we acknowledge that the execution efficiency of the framework is indeed influenced by the inference speed of the base model. In order to address this, we have conducted performance evaluations and execution time measurements of our framework on various base models including Zephyr and GLM. The EL results are shown in Table~\ref{table:main}, while the running time analysis is illustrated in Table~\ref{table:runtime}. The execution time is obtained by randomly sampling 100 tests on each dataset without any parallelization acceleration. 

\begin{table*}[!t]
\centering
\resizebox{\textwidth}{!}{
\begin{threeparttable}
\begin{tabular}{c c|c c|c c|c c|c c} \toprule
\multicolumn{2}{c|}{\textbf{Domain}} & \multicolumn{2}{c|}{\textbf{Forgotten Realms}} & \multicolumn{2}{c|}{\textbf{Lego}} & \multicolumn{2}{c|}{\textbf{Star Trek}} & \multicolumn{2}{c}{\textbf{Yugioh}}  \\ \midrule
\multicolumn{2}{c|}{Method} & N.Acc. & U.Acc. & N.Acc. & U.Acc.& N.Acc. & U.Acc.& N.Acc. & U.Acc.\\ \midrule
\multirow{2}{*}{Tradition}&Blink & \textbf{0.590} & 0.208 & 0.456 & 0.240 & 0.371 & 0.080 & 0.377 & 0.132 \\ 
&MetaBlink & 0.563 & 0.391 & 0.507 & 0.396 & 0.346 & 0.213 & 0.380 & 0.233 \\
\midrule
\multirow{2}{*}{LLM}&Llama3(Text) & 0.095 & 0.079 & 0.040 & 0.033 & 0.012 & 0.008 & 0.009 & 0.005 \\ 
&Llama3(Sum.) & 0.235 & 0.196 & 0.200 & 0.163 & 0.150 & 0.099 & 0.054 & 0.033 \\ 
\midrule
Ours&OneNet & 0.558 & \textbf{0.465} & \textbf{0.538} & \textbf{0.437} & \textbf{0.539} & \textbf{0.355} & \textbf{0.408} & \textbf{0.248} \\ \bottomrule
\end{tabular}
\end{threeparttable}
}
\caption{Normalized and Unnormalized Accuracy on ZeShEL Dataset in Different Domains}
\label{table:ZeShEL}
\end{table*}

Overall, while the GLM model shows slightly lower performance than Zephyr, its inference speed is up to ten times faster. To assess the cost of our framework, Table~\ref{table:input_tokens} Appendix~\ref{Ap:exp} presents the average input tokens per module, indicating that our framework minimally increases token usage compared to direct LLM application. Additionally, some modules, such as ERS, can run offline, further improving efficiency. As noted in the Limitation section, ongoing research on optimizing LLM inference speed is expected to further enhance the efficiency of our framework, which we believe can further enhance the efficiency of our framework.

\section{Framework Generalization} \label{Ap:general}

As discussed in Section~\ref{Intro}, traditional methods heavily depend on external data such as entity priors, preventing their adaptability across different scenarios~\cite{mulrel18}. Moreover, these methods struggle with practical issues such as entity ID mapping, further complicating their migration across various knowledge bases~\cite{ner4el21}. In contrast, our framework, illustrated in Figure~\ref{fig:framework}, leverages large language model textual reasoning and requires no fine-tuning, which enables our model to perform entity linking across diverse domains and knowledge bases.

To further discuss the generalizability of our framework, we present the normalized performance on the ZeShEL~\cite{logeswaran-etal-2019-zeshel} dataset in Table~\ref{table:ZeShEL}. ZeShEL is an entity linking dataset constructed using Wikias from Fandom. We compare the performance of OneNet with Blink~\cite{blink20}, MetaBlink~\cite{li2022metablink} and the base model in few-shot setting. We used the settings described in Section~\ref{benchmark}. However, due to the excessive length of text in ZeshEL, the base model with original entity text resulted in poor performance. To address this, we also report the base model performance with LLM-generated entity summary, which reduces the context length. 

In Table~\ref{table:ZeShEL}, our OneNet achieves optimal results in most domains, demonstrating the effectiveness of our approach. Note that OneNet improves about 35\% compared to a single LLM, which further prove the generalization of our framework. 
However, a performance gap remains compared to the results reported in Table~\ref{table:main} for the wiki-based dataset. We attribute this disparity partially to the influence of non-wiki data, but more significantly to the excessive length of ZeShEL's text. As shown in Table~\ref{table:statistics}, ZeShEL surpasses other datasets in terms of the number of candidates, entity descriptions, and contexts, especially the contexts. The irrelevant information in excessively long contexts can mislead LLMs~\cite{shi2023longcontext}. To address this, we propose extracting critical information from contexts, such as the first and last sentences of paragraphs and sentences containing mentions, to enhance the performance of our framework.

% \begin{table}[!t]
% \centering
% \resizebox{\linewidth}{!}{
% \begin{tabular}{c|c c c c c c c } \toprule
% \textbf{Method} & \textbf{ACE2004} & \textbf{AIDA} & \textbf{AQUAINT} & \textbf{CWEB} & \textbf{MSNBC} & \textbf{WIKI} & \textbf{ZeShEL} \\ \midrule
% Cand. Nums	& 42.25	& 7.18	& 34.69	& 6.98	& 25.93	& 6.09 & \textbf{55.20}\\
% Ent. Tokens	& 190.79	& 262.16	& 197.00	& 239.36	& 198.81	& 227.04 & \textbf{441.65}\\
% Cont. Tokens& 171.17	& 452.53	& 169.57	& 222.126	& 198.88	& 195.90 & \textbf{2394.93}\\ \bottomrule
% \end{tabular}}
% \caption{Average of Candidates Numbers, Entity Description Tokens, and Context Tokens across Datasets.}
% \label{table:ZeShEL Result}
% \end{table}

\section{Prompt} \label{Ap:prompt}
In order to understand more intuitively how we prompted the different modules, Tables~\ref{table:module_prompt_example} and~\ref{table:instruction_prompt_example} show example prompts for all the modules and the specific context of the entity link instructions prompts respectively. Specifically, most of the prompts in Table~\ref{table:module_prompt_example} were written by hand to achieve the functionality we wanted, while most of the prompts in Table~\ref{table:instruction_prompt_example} were generated by GPT to distill knowledge from stronger models.

% \newpage

\begin{table*}[htbp] 
\begin{tabular}{c p{13cm}c} \hline
Module & \qquad \qquad \qquad \qquad \qquad \qquad \qquad \qquad Prompt \\ \hline
Summarization & The following is a description of \{mention\}. Please extract the key information of \{mention\} and summarize it in one sentence: \{description\} \\ \hline
Point-wise EL & You're an entity disambiguator. I'll give you the description of entity disambiguation and some tips on entity disambiguation, and you need to pay attention to these textual features: \{Instruction Prompt\}. Now, I'll give you a mention, a context, and a candidate entity, and the mention will be highlighted with '\#\#\#'. Mention:\{mention\}, Context:\{context\}, Candidate Entity:{one candidate entity}. You need to determine if the mention and the candidate entity are related. Please refer to the above tips and give your reasons, and finally answer 'yes' or 'no'. Answer 'yes' when you think the information is insufficient or uncertain. \\ \hline
Category & You are a mention classifier. Wikipedia categorizes entity into the following categories: {Categories}. Now, I will give you a mention and its context, the mention will be highlighted with '\#\#\#'. Mention:\{mention\}, Context:\{context\}. please determine which of the above categories the mention {mention} belongs to? \\ \hline
Contextual EL & You're an entity disambiguator. I'll give you the description of entity disambiguation and some tips on entity disambiguation, you should pay attention to these textual features: \{Instruction Prompt\}. The following example will help you understand the task: \{CoT Prompt\}. Now, I'll give you a mention, a context, and a candidate entity, and the mention will be highlighted with '\#\#\#'. Mention:\{mention\}, Context:\{context\}, \{Candidates\} .You need to determine which candidate entity is more likely to be the mention. Please refer to the above example, give your reasons, and finally answer serial number of the entity and the name of the entity. If all candidate entities are not appropriate, you can answer '-1.None'.\\ \hline
Prior EL & You're an entity disambiguator. I'll provide you a mention and its candidates below. mention:\{mention\}. \{Candidates\}. Based on your knowledge, please determine which is the most likely entity when people refer to mention "\{mention\}", and finally answer the name of the entity.\\ \hline
Merge & You're an entity disambiguator. I'll give you the description of entity disambiguation and some tips on entity disambiguation, you should pay attention to these textual features: \{Instruction Prompt\}. Now, I'll give you a mention, a context, and a candidate entity, and the mention will be highlighted with '\#\#\#'. Mention:\{mention\}, Context:\{context\}, \{Candidates\} .You need to determine which candidate entity is more likely to be the mention. Please refer to the above example, give your reasons, and finally answer serial number of the entity and the name of the entity. If all candidate entities are not appropriate, you can answer '-1.None'. \\ \hline
\end{tabular}
\caption{Examples of Prompt for Each Module}
\label{table:module_prompt_example}
\end{table*}

\newpage

\begin{table*}[htbp] 
\begin{tabular}{c p{14cm}c} \hline
Id & \qquad \qquad \qquad \qquad \qquad \qquad \qquad \qquad Context \\ \hline
Prompt 0 & Entity Disambiguation Task: You will be given a context, a mention, and a set of candidate entities. Your goal is to identify the entity that corresponds to the mention within the context. Follow these steps:1. Read the context and identify the mention.2. Examine the candidate entities provided.3. Consider the following features to determine the best match for the mention:a. Categories: Look for category labels or descriptions that align with the mention.b. Modifiers: Pay attention to qualifying words that provide additional information about the mention.c. Contextual clues: Analyze the surrounding text for related entities, events, or relationships.d. Semantic meaning: Consider the meaning, context, and purpose of the mention and candidate entities. 4. Make an informed decision based on the available information and select the most suitable entity. \\ \hline
Prompt 1 & Here's a hint to help your friend understand entity disambiguation and some features to consider:Entity disambiguation involves determining if a given candidate entity is the same as the mention within a given context. To make an accurate judgment, consider the following features:1. Categories: Look for clues indicating the category or type of the mention and the candidate entity. Are they both people, places, organizations, or something else? Matching categories often indicate a higher likelihood of being the same entity.2. Modifiers: Pay attention to descriptive words or phrases that modify the mention and the candidate entity. Do they share similar modifiers? For example, if the mention is 'red apple' and the candidate entity is 'juicy apple,' the shared modifier 'apple' suggests a potential match. 3. Contextual information: Analyze the surrounding text to understand the context in which the mention and candidate entity appear. Look for additional information that can help determine if they refer to the same entity. Consider factors such as location, time, relationships, or events mentioned. 4. Unique identifiers: Check for any unique identifiers associated with the mention and the candidate entity. These could be specific names, titles, dates, or other distinct attributes. Matching unique identifiers can strongly indicate a match. 5. Disambiguation cues: Look for disambiguation cues within the context that explicitly clarify or distinguish between different entities. These cues may include pronouns, definite or indefinite articles, or explicit references to other entities. Remember, entity disambiguation can sometimes be challenging, especially when dealing with ambiguous or incomplete information. It's important to carefully analyze the given context and consider multiple features to make an informed decision. \\ \hline
\end{tabular}
\end{table*}

\newpage

\begin{table*}[htbp]
\begin{tabular}{c p{14cm}c} \hline
Id & \qquad \qquad \qquad \qquad \qquad \qquad \qquad \qquad Context \\ \hline 
  Prompt 2 & Entity disambiguation is a common task in natural language processing (NLP) and information retrieval. The goal is to determine which specific entity is being referred to in a text when there may be multiple entities with the same or similar names. Here are some hints and features to look out for when you're doing this task manually: Context: The surrounding sentence or paragraph where the mention is located can provide clues about the entity. For example, if the mention is 'Apple' and the context is about technology or smartphones, it's likely referring to the technology company. If the context is about fruit or food, it's probably referring to the fruit. Categories: Entities often belong to specific categories or types, such as people, organizations, locations, etc. If you know the category of the candidate entity, this can help you decide if it matches the mention. For example, if the mention is 'Washington' and the candidate entity is a person (e.g., George Washington), but the context is about places, then the candidate entity is probably not a match. Modifiers: These are words or phrases that modify or add details to the mention. For example, in the mention 'President Obama,' the modifier 'President' indicates that the entity is a person, specifically Barack Obama. Modifiers can also include adjectives, descriptive phrases, or other context that helps specify the entity. Co-references: These are other mentions of the same entity in the text. If the text refers to 'Apple' multiple times and talks about both smartphones and fruit, you might be able to determine which 'Apple' is being referred to based on how it's discussed elsewhere in the text.  Temporal and Geographical Factors: The time and place that the text was written can also provide clues. For example, if the mention is 'Jordan' in an article written in the 1990s about basketball, it's likely referring to Michael Jordan. If it's in a recent article about Middle Eastern politics, it's probably referring to the country Jordan. External Knowledge: Sometimes, you might need to use knowledge that's not contained in the text. For example, if the mention is 'Musk' and the context is about space travel, you might need to know that Elon Musk is the CEO of SpaceX to realize that 'Musk' refers to him. Remember, entity disambiguation can be tricky, and there might not always be a clear answer. It often requires a combination of understanding the text, knowing about the world, and using your best judgment. \\ \hline
Prompt 3 & Context: Look at the surrounding text to understand the topic.Categories: Consider the type of the entity (person, organization, location, etc.).Modifiers: Pay attention to words or phrases that add details to the mention.Co-references: Check other mentions of the same entity in the text.Temporal and Geographical Factors: Consider when and where the text was written.External Knowledge: Use outside knowledge not contained in the text.Remember, entity disambiguation requires understanding the text, knowing about the world, and using good judgment. \\ \hline

\end{tabular}
\end{table*}

\begin{table*}[htbp] 
\begin{tabular}{c p{14cm}c} \hline
Id & \qquad \qquad \qquad \qquad \qquad \qquad \qquad \qquad Context \\ \hline
Prompt 4 & **Entity Disambiguation Task** Objective: Your goal is to identify the correct entity from a list of candidate entities that corresponds to a given mention within a specific context. Procedure: You will be provided with three things: 1. Context: This is a paragraph or a set of sentences that provides the surrounding information where the mention is found. 2. Mention: This is the specific term or phrase that you need to disambiguate – i.e., to identify its correct meaning or reference. 3. Candidate Entities: This is a list of possible entities that the mention could refer to. Your job is to select the correct one based on the context. Features to Look Out For: 1. **Categories/Types**: Entities belong to different categories such as people, organizations, locations, events, etc. The category of the mention can often be inferred from the context. For instance, if the context is discussing a concert, the mention is likely referring to a musician or a music-related entity.  2. **Modifiers**: These are words or phrases that provide additional information about the mention. For example, in the mention 'Apple CEO Tim Cook', 'Apple CEO' is a modifier that helps distinguish this Tim Cook from other individuals with the same name. 3. **Co-references**: These are other mentions of the same entity in the context. They can provide additional clues about the entity. For example, if the context mentions 'the tech giant' before mentioning 'Apple', these two are co-references pointing to the same entity. 4. **Temporal and Spatial Clues**: The time and place mentioned in the context can also help in disambiguating the entity. For example, if the context is about the 19th century, a mention of 'Washington' is more likely to refer to George Washington than the city of Washington D.C. 5. **Domain-specific Knowledge**: Sometimes, general world knowledge or domain-specific knowledge can help disambiguate entities. For example, if the context is about computer programming, a mention of 'Python' is likely referring to the programming language, not the snake. Remember, the goal is to use the context and your understanding of the world to determine which entity from the list of candidates the mention is most likely referring to. It's not always easy, and there may be times when more than one candidate seems possible. In such cases, choose the one that seems most likely based on all the available information. Good luck! \\ \hline
Prompt 5 & **Entity Disambiguation Task** Goal: Identify the correct entity from a list of candidates that matches a given mention within its context.Procedure: You'll get a context (surrounding text), a mention (term to identify), and candidate entities (possible matches). Key Features: 1. Categories: Check if the context implies a category (person, place, etc.) for the mention. 2. Modifiers: Look for additional info (e.g., 'Apple CEO Tim Cook') that distinguishes the mention.3. Co-references: Find other mentions of the same entity in the context for extra clues.4. Temporal/Spatial Clues: Time and place details can help disambiguate the entity.5. Domain Knowledge: Use general or specific knowledge to infer the correct entity. Use all available information to select the most likely entity from the candidates. Good luck! \\ \hline
\end{tabular}
\caption{The Instruction Prompts of Entity Linking Generated by GPT}
\label{table:instruction_prompt_example}
\end{table*}

\end{document}